\documentclass[conference]{IEEEtran}
\usepackage[utf8]{inputenc}
\usepackage[T1]{fontenc}
\IEEEoverridecommandlockouts
\usepackage{cite}
\usepackage{amsmath,amssymb,amsfonts}
\usepackage{algorithmic}
\usepackage{graphicx}
\usepackage{textcomp}
\usepackage{multicol}
\usepackage{multirow}
\usepackage{xcolor}
\usepackage{subfigure}
\def\BibTeX{{\rm B\kern-.05em{\sc i\kern-.025em b}\kern-.08em
    T\kern-.1667em\lower.7ex\hbox{E}\kern-.125emX}}
\begin{document}

\title{Periodicity-Enforced Neural Network for Designing Deterministic Lateral Displacement Devices}

\author{\IEEEauthorblockN{Andrew Lee}
\IEEEauthorblockA{\textit{School of EECS} \\
\textit{Pennsylvania State University}\\
University Park, PA, USA \\
ayl5530@psu.edu}
\and
\IEEEauthorblockN{Mahir Mobarrat}
\IEEEauthorblockA{\textit{School of Engineering} \\
\textit{Rensselaer 
Polytechnic Institute}\\
Troy, NY, USA \\
mobarm@rpi.edu}
\and
\IEEEauthorblockN{Xiaolin Chen}
\IEEEauthorblockA{\textit{School of ENCS} \\
\textit{Washington State University Vancouver}\\
Vancouver, WA, USA \\
chenx@wsu.edu}
}

\maketitle


\begin{abstract}
Deterministic Lateral Displacement (DLD) devices enable liquid biopsy for cancer detection by separating circulating tumor cells (CTCs) from blood samples based on size, but designing these microfluidic devices requires computationally expensive Navier-Stokes simulations and particle-tracing analyses. While recent surrogate modeling approaches using deep learning have accelerated this process, they often inadequately handle the critical periodic boundary conditions of DLD unit cells, leading to cumulative errors in multi-unit device predictions. This paper introduces a periodicity-enforced surrogate modeling approach that incorporates periodic layers—neural network components that guarantee exact periodicity without penalty terms or output modifications—into deep learning architectures for DLD device design. The proposed method employs three sub-networks to predict steady-state, non-dimensional velocity and pressure fields (u, v, p) rather than directly predicting critical diameters or particle trajectories, enabling complete flow field characterization and enhanced design flexibility. Periodic layers ensure exact matching of flow variables across unit cell boundaries through architectural enforcement rather than soft penalty-based approaches. Validation on 120 CFD-generated geometries demonstrates that the periodic layer implementation achieves 0.478\% critical diameter error while maintaining perfect periodicity consistency, representing an 85.4\% improvement over baseline methods. The approach enables efficient and accurate DLD device design with guaranteed boundary condition satisfaction for multi-unit device applications.
\end{abstract}



\section{Introduction}

Cancer remains one of the leading causes of mortality worldwide, resulting in over 600,000 fatalities only in United States \cite{Sung2021}. As a result, early detection is crucial for improving patient outcomes and survival rates. Traditional cancer diagnostics have relied on solid biopsies obtained from surgically removed tumor tissues. This approach faces significant limitations: tumors must be in sufficient size for imaging detection, the invasive nature causes patient discomfort and complication, and tissue accessibility may limit the early diagnosis opportunities. Consequently, researchers have increasingly turned to liquid biopsy, which identifies circulating tumor cells (CTCs), cell-free DNA (cfDNA), RNA, or other biomarkers within blood samples \cite{wan2017liquid}.

Among liquid biopsy technologies, CTC separation represents critical enablement for downstream analysis and characterization. Detection of CTCs offered early cancer detection through minimally invasive procedure. Moreover, detection through blood samples, unlike traditional tissue biopsies, enables real-time monitoring of tumor progression and further analysis of the treatment results \cite{TLCR73085} \cite{Crowley2013} \cite{Allen2024}.

Various microfluidic techniques have emerged, such as immunoaffinity capture, dielectrophoresis, and acoustic separation. However, Deterministic Lateral Displacement (DLD) devices represent a promising microfluidic technology for liquid biopsy applications with their unique advantages: size-based, label-free separation of particles, requiring no antibodies, fluorescent dyes, or magnetic beads. These devices consist of arrays of posts that separate particles based on size through hydrodynamic forces. DLD devices achieve particle separation through separation based on a critical diameter ($D_c$): particles smaller than $D_c$ follow streamlines in "zig-zag mode," while particles larger than $D_c$ are displaced laterally in "bumped mode." This size-based separation mechanism makes DLD devices particularly suitable for isolating CTCs from blood components \cite{mcgrath2014deterministic}.

The central design challenge in DLD devices lies in precisely tuning the critical diameter to match target cell populations. CTCs typically range from 15-25 $\mu m$ in diameter, significantly larger than most blood cells, but this size difference varies across cancer types and patient populations. The critical diameter depends on complex interactions between device geometry (post diameter, gap size, array tilt) and flow conditions, requiring precise control for effective separation. Conventional design approaches rely on Computational Fluid Dynamics (CFD) simulations of the Navier-Stokes equations, which create substantial computational bottlenecks. However, CFD methods require extensive compute time that often limits the size of microfluidic devices that can be simulated. Designing with CFD requires repetitive 2-3 hours of simulations for every new geometry for testing \cite{mahir_thesis}.

To address these computational bottlenecks, recent studies have introduced surrogate modeling using deep learning techniques. Vatandoust et al. developed the DLDNN framework, which combines convolutional neural networks (CNNs) and fully connected neural networks to map device geometries to velocity fields and predict $D_c$ \cite{vatandoust2022dldnndeterministiclateraldisplacement}. Alternative approaches train classifiers or regressors on simulated particle trajectories \cite{chen2024poster}. While these surrogates have successfully reduced computation time by several orders of magnitude, they face a critical limitation in boundary condition enforcement.

The fundamental challenge lies in periodic boundary conditions. DLD devices consist of repeating unit cells, and accurate multi-unit device modeling requires exact matching of flow variables (velocity and pressure) across unit cell boundaries. Existing approaches approximate periodicity through penalty terms in loss functions or output modifications, which cannot guarantee exact boundary matching. These approximations as small as 1\% error may seem acceptable for single-unit predictions, but errors accumulate across multiple unit cells in practical DLD devices, potentially leading to significant trajectory prediction errors for CTCs, ending up with significant errors when biopsy passes 50 to 100 repeated units.

Recent advances in periodic neural network layers offer a solution to this boundary condition problem. Dong et al. introduced periodic layers of neural networks that map inputs through smooth periodic functions, ensuring outputs automatically satisfy periodic boundary conditions without penalty terms \cite{Dong_2021}. These layers guarantee exact periodicity by construction, providing architectural enforcement rather than optimization-based approximation.

This paper develops from approaches from Dong et al. to apply into DLD device's vertical periodicity and proposes a periodicity-enforced neural network architecture that integrates these periodic layers into deep learning surrogates for DLD device design. Unlike prior methods that approximate periodicity with loss penalties, our approach guarantees exact periodicity through architectural modification of the neural network structure. We extend prediction capabilities to full velocity and pressure fields ($u, v, p$) rather than direct critical diameter estimation, enabling both accurate critical diameter computation and comprehensive flow analysis for diverse cancer cell sizes. Validation on CFD-generated datasets demonstrates that periodic layer integration achieves superior accuracy while guaranteeing exact periodicity, providing a robust foundation for reliable multi-unit DLD device modeling.


\section{DLD Device}

\subsection{Geometry}
\begin{figure}
    \centering
    \includegraphics[width=0.5\linewidth]{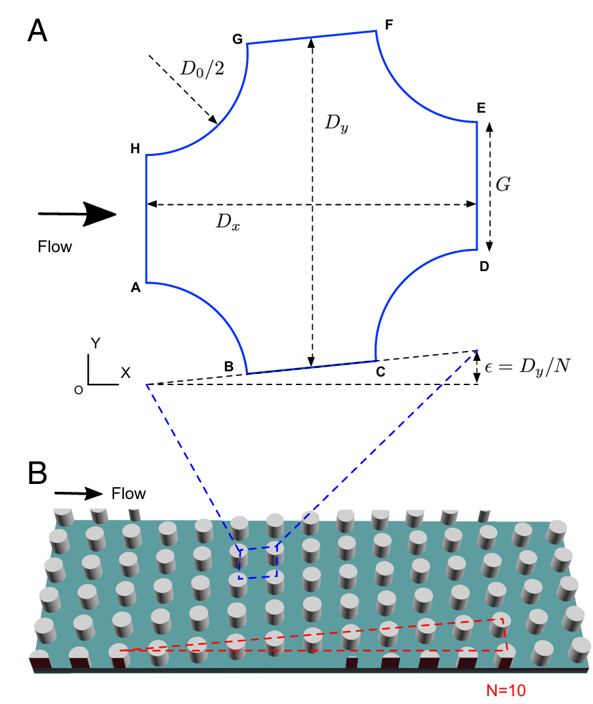}
    \caption{Geometry Example of A) Single Unit Cell of DLD Device B) Full Device \cite{DLDIMAGE}}
    \label{fig:DLDGeometry}
\end{figure}

DLD device unit cells consist of four cylindrical posts positioned at corners, each with diameter $D_0$, as shown in Fig.~\ref{fig:DLDGeometry}. Each unit cell has horizontal length $D_x$ and vertical height $D_y$, where both dimensions equal $D_s$ (set to 0.4 for training). The horizontal domain spans 0 to 0.4, while the vertical domain depends on the tilt angle $\epsilon$, defined as:
\begin{equation}
    \epsilon = \frac{D_s}{N}
\end{equation}
where $N$ represents the period number of the post array.

\subsection{Particle Separation Through Bumped or Zig-zag Mode}

Particle separation in DLD devices relies on the hydrodynamic interaction between flowing particles and the post array geometry. The separation mechanism is governed by the critical diameter $D_c$, which represents a threshold size that determines particle behavior within the device.

For particles smaller than the critical diameter ($D_p < D_c$), the hydrodynamic forces are insufficient to overcome the streamline curvature around posts. These particles follow the fluid streamlines in a "zig-zag mode," oscillating between posts while maintaining their overall flow direction parallel to the main channel. The particle trajectory mirrors the streamline pattern, resulting in minimal lateral displacement across the device width.

Conversely, particles larger than the critical diameter ($D_p > D_c$) experience stronger hydrodynamic interactions that force them away from streamlines. These particles enter "bumped mode," where they are systematically displaced laterally with each post encounter. The cumulative effect of multiple collisions with the post array creates a net lateral migration across the device width, effectively separating large particles from the main flow stream.

This size-based separation mechanism makes DLD devices particularly effective for CTC isolation applications. CTCs, typically ranging from 12-25 $\mu m$ in diameter, are significantly larger than most blood cells (red blood cells:  $\sim$8 $\mu m$, white blood cells: 5-20 $\mu m$) \cite{HAO20183}. By designing the critical diameter to fall between these size ranges, DLD devices can selectively capture CTCs in bumped mode while allowing smaller blood components to pass through in zig-zag mode. While some larger white blood cells may also displaced laterally with bumped mode, the output sample will be enriched with CTCs, sufficient for cancer detection.


\section{Related Works}

\subsection{Surrogate Modeling of DLD Devices}

Machine learning approaches for accelerating DLD device design have emerged along two primary paradigms: direct trajectory prediction and flow-field prediction. 

Chen et al. employed direct trajectory prediction, where geometric parameters serve as inputs to models that directly output particle coordinates as time series \cite{chen2024poster}. However, this approach faces substantial challenges: models must train on complete DLD device geometries requiring enormous datasets and extensive training time, undermining the acceleration goal. Additionally, time-series outputs introduce temporal dependencies and non-smooth predictions due to discretization requirements.

In contrast, Vatandoust et al. developed DLDNN, a flow-field prediction framework utilizing a two-stage architecture where a CNN predicts velocity fields from device geometries, followed by an FCNN that estimates $D_c$ \cite{vatandoust2022dldnndeterministiclateraldisplacement}. This approach operates on single unit cells rather than full devices, reducing training data requirements and computational time by several orders of magnitude. Our work adopts this flow-field prediction paradigm for its computational efficiency and scalability.

However, existing approaches exhibit a critical limitation in boundary condition enforcement. Current methods approximate periodicity through penalty terms in loss functions rather than architectural guarantees. While these soft enforcement approaches may achieve acceptable accuracy for single-unit predictions, they cannot guarantee exact boundary matching required for multi-unit device modeling. This limitation becomes particularly problematic when scaling predictions across multiple unit cells, where small boundary mismatches can accumulate into significant trajectory prediction errors.

\subsection{Periodic Neural Network Layers}

Dong et al. introduced periodic layers that map inputs through smooth periodic functions, ensuring outputs automatically satisfy periodic boundary conditions without penalty terms or output modifications \cite{Dong_2021}. These layers guarantee exact periodicity by construction through architectural design rather than optimization-based approximation. 

For neural networks processing spatial data with periodic boundaries, this approach provides a fundamental advantage over traditional methods. Instead of relying on loss function penalties that compete with primary training objectives, periodic layers enforce constraints through the network structure itself. This architectural solution eliminates the possibility of boundary condition violations that plague optimization-based approaches.

The periodic transformation typically involves trigonometric functions applied to spatial coordinates, ensuring that any function computed using these transformed inputs maintains strict periodicity. For DLD devices with inherently periodic unit cell structures, this approach provides an ideal solution to boundary condition enforcement challenges.

\subsection{Physics-Informed Neural Networks (PINNs)}

Physics-Informed Neural Networks represent an alternative paradigm that embeds governing equations directly into training objectives by minimizing PDE residuals alongside boundary conditions \cite{RAISSI2019686}. This approach enables fluid dynamics solutions without labeled data by incorporating physics knowledge into the learning process.

While PINNs offer the advantage of eliminating dataset requirements, they face challenges in complex geometries with multiple boundary conditions. Proper enforcement of periodic boundaries, wall conditions, and inlet/outlet specifications creates complex multi-constraint optimization problems \cite{Sun_2020}. The integration of periodic layers with PINN approaches represents a promising direction for future research, potentially combining the benefits of exact boundary condition enforcement with physics-driven training.

However, the current work focuses on improving existing supervised surrogate modeling through architectural enhancements, leaving physics-informed approaches for future investigation. The demonstrated effectiveness of periodic layers in supervised settings provides a foundation for potential PINN integration in subsequent research.


\section{Methodology}

\subsection{Dataset Generation}

CFD simulations generate 120 geometries with varying post diameter ($D_0$) and period number ($N$). Post diameters are characterized by the non-dimensional parameter:
\begin{equation}
    F = \frac{D_0}{D_s}
\end{equation}
where $F$ ranges from 0.05 to 0.70 in increments of 0.05, and $N$ varies from 3 to 14. 

\begin{equation}
    Re = \frac{\rho U L}{\mu}
    \label{Re}
\end{equation}

The Reynolds number ($Re$) is defined as in Eqn.~\ref{Re}, where $\rho$ is fluid density, $U$ is characteristic velocity, $L$ is characteristic length (gap size: $D_s - D_0$), and $\mu$ is dynamic viscosity. The Reynolds number the ratio of inertial forces to viscous forces in a fluid flow, determining whether the flow will be laminar or turbulent. In this paper, $Re$ varies across geometries in the CFD simulations but is not included as a training input parameter, focusing the model on geometric effects only.

The pressure-based inlet dataset is used where pressure difference ($\Delta p$) is specified:
\begin{equation}
    \Delta p = p_{inlet} - p_{outlet}
    \label{dp}
\end{equation}
The flow rate and velocity profile develop naturally from the geometry and let the network determine the inlet velocity distribution. This approach allows more physically realistic modeling for sudden contractions or expansion in comparison to parabolic inlet. 

The parabolic inlet: 
\begin{equation}
    u(0,y) = u_{max}\left[1 - \left(\frac{y}{h}\right)^2\right]
    \label{parabolic_inlet}
\end{equation}
where $h$ represents half-channel height which is, $h$ = $\frac{D_s}{2}$ = 0.2. The parabolic inlet assumes fully developed laminar flow entering the device, which is only appropriate when inlet channel is long enough for flow development.

Each dataset entry contains $(x, y)$ coordinates, geometric parameters $(F, N)$, and target fields $(u, v, p)$. Approximately 1,000 randomly sampled coordinate points per geometry yield 120,000 total samples.

\subsection{Network Architecture}

\begin{figure}
    \centering
    \includegraphics[width=\linewidth]{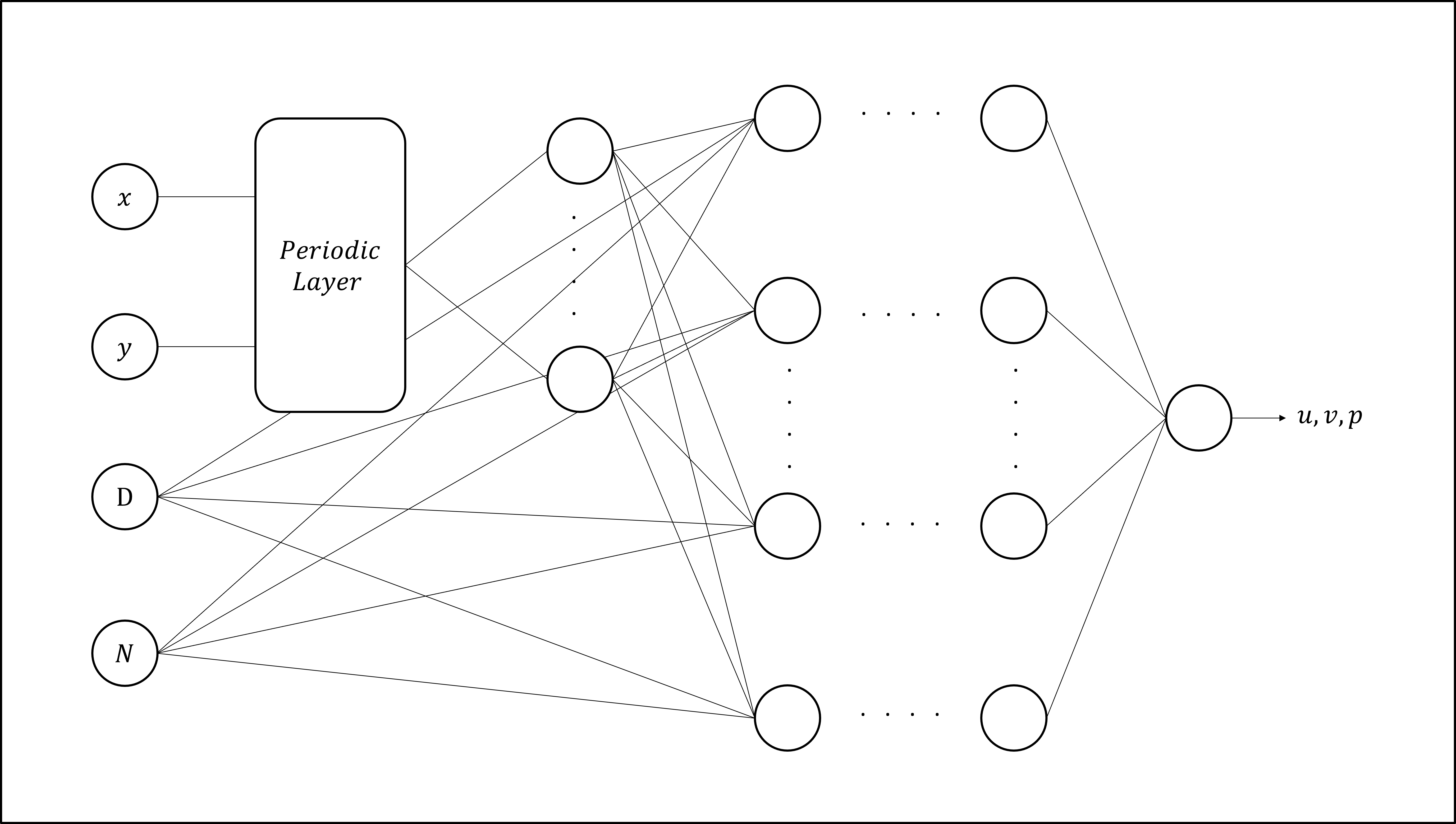}
    \caption{Network Architecture with Periodic Layer}
    \label{fig:PeriodicNN}
\end{figure}

The baseline architecture follows Sun et al., employing 50 neurons across three layers with Swish activation for smooth differentiability with loss term modified for soft boundary enforcements by past study by Mobarrat \cite{Sun_2020} \cite{mahir_thesis}. Our proposed architecture (Fig.~\ref{fig:PeriodicNN}) uses 64 neurons across 8 hidden layers, with the first layer implementing periodic transformation. The periodic layer processes spatial coordinates $(x, y)$ using:

\begin{equation}
    y_{periodic} = \sin(2\pi y)
\end{equation}
ensuring strict periodicity in the y-direction for DLD boundary conditions.

The periodic layer outputs 64 neurons that concatenate with geometric parameters $(D_0, N)$ to form inputs for subsequent layers. While standard hidden layers use Swish activation, the periodic layer employs tanh activation for zero-centered, bounded outputs that properly capture periodic behavior. Three separate sub-networks predict velocity components $(u, v)$ and pressure $(p)$, each outputting single scalar values.

\subsection{Training Procedure}

The model trains using Adam optimizer with initial learning rate $10^{-3}$, decaying by 0.5 every 50 epochs over 1,000 epochs with batch size 2,000. The loss function combines dataset and boundary components:

\textbf{-Dataset Loss:}
\begin{equation}
    L_{data} = \frac{1}{n}\sum_{i=1}^{n}[(u_i-\hat{u_i})^2 + (v_i-\hat{v_i})^2 + (p_i-\hat{p_i})^2]
\end{equation}
where $n$ is sample count, $(u,v,p)$ are CFD-computed fields, and $(\hat{u},\hat{v},\hat{p})$ are predictions.

\textbf{-Boundary Conditions (Soft Enforcement):}
\begin{itemize}
    \item Zero velocity at walls: $u = v = 0$
    \item Zero vertical velocity at inlet: $v = 0$
    \item Fixed pressure drop at inlet: $p = \Delta p = 0.1$
\end{itemize}

Training requires approximately 6 hours on NVIDIA RTX 4060 Laptop with 8GB VRAM.

\subsection{Critical Diameter Calculation}

Critical diameters are computed by integrating particle streamlines from initial positions across unit cells using predicted velocity fields. For the trajectory calculation, we employ a massless particle model where particles with negligible inertia are assumed to perfectly follow fluid streamlines. The particle motion is governed by:
\begin{equation}
    \frac{d\mathbf{x}_p}{dt} = \mathbf{v}_f(\mathbf{x}_p)
\end{equation}
where $\mathbf{x}_p$ is the particle position and $\mathbf{v}_f(\mathbf{x}_p)$ is the fluid velocity field interpolated at the particle's location. This simplification is valid for microscale flows where particle inertia is negligible compared to viscous forces. When particles encounter solid boundaries (post walls), their trajectories are altered based on a reflection principle where the particle retains its tangential velocity while its normal velocity is reversed.

For each geometry, $D_c$ represents the smallest particle diameter exhibiting bumped mode behavior. Predicted $D_c$ values are compared against CFD-derived references using relative percent error.

\subsection{Evaluation}

Additional validation dataset is generated from geometric constants outside of F domain of training dataset. The dataset includes 144 geometries for validation. Meanwhile, the evaluation is done on both training and validation datasets to verify the model's effectiveness on fitting training data and generalizing to predict validation datasets. 

The metrics used for evaluating performance are R² score of the flow field prediction and the critical diameter percent error. The R² score is calculated as:
\begin{equation}
    R^2 = 1 - \frac{\sum_{i=1}^{n}(y_i - \hat{y_i})^2}{\sum_{i=1}^{n}(y_i - \bar{y})^2}
\end{equation}
where $y_i$ represents the CFD-computed values, $\hat{y_i}$ represents the predicted values, $\bar{y}$ is the mean of CFD values, and $n$ is the number of data points. The higher R² score with the maximum of 1 represents better model fit and accuracy.

The critical diameter percent error is computed as:
\begin{equation}
    \text{Percent Error} = \frac{|D_{c,true} - D_{c,pred}|}{|D_{c,true}|} \times 100\%
\end{equation}
where $D_{c,true}$ is the CFD-derived critical diameter and $D_{c,pred}$ is the model-predicted critical diameter.

These metrics are used to determine the performance of the model, while field error plots also visually verify capturing flow patterns and magnitude across the unit. The lower percent error represents better accuracy with the minimum of 0.


\section{Results}

\subsection{Field Predictions}

\begin{figure*}
    \centering
    \includegraphics[width=0.70\textwidth]{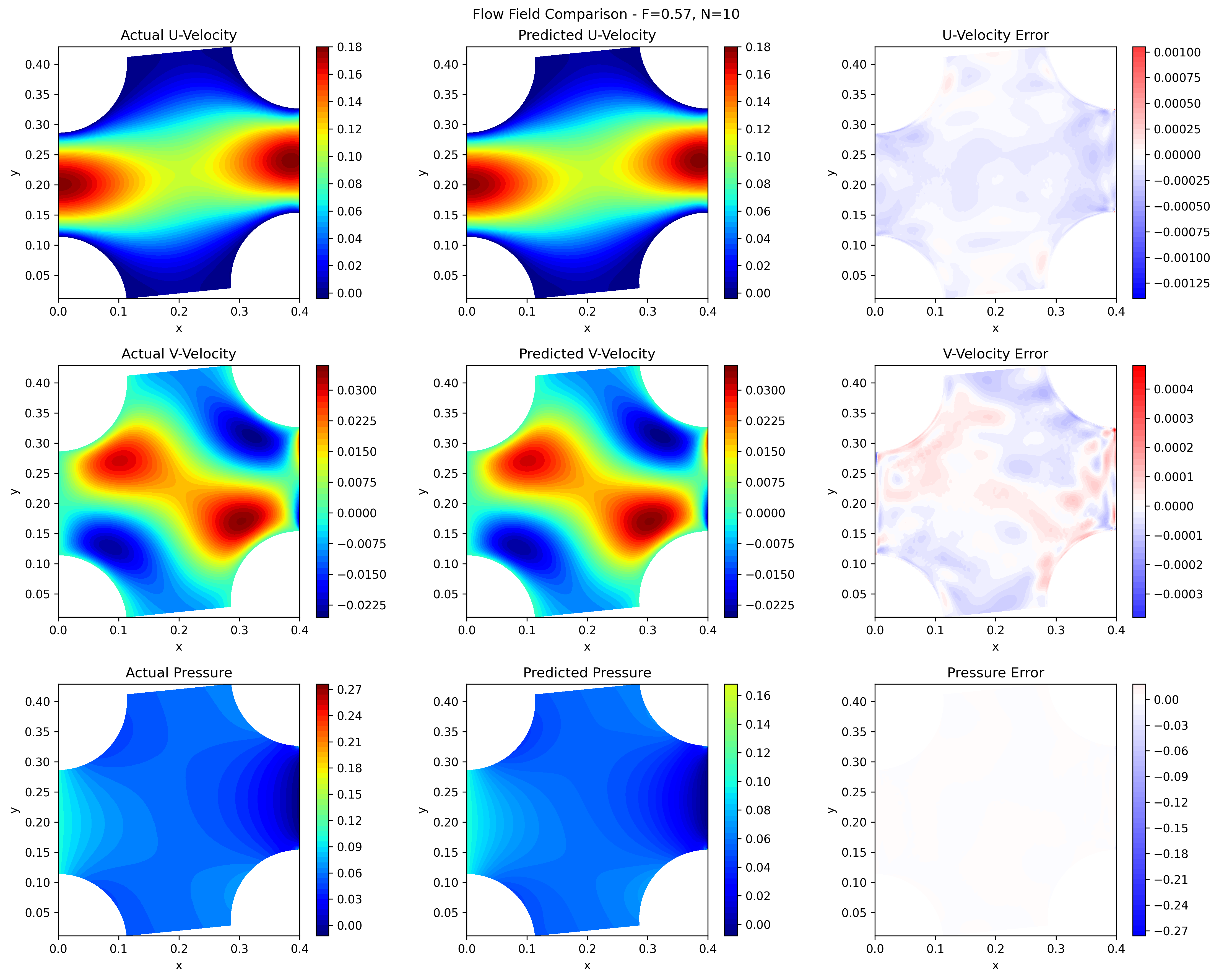}
    \caption{Field Prediction Results in Comparison to CFD Simulations for Proposed Model From Validation Data of F = 0.57 and N = 10}
    \label{fig:FieldPred_sup_F0_57_N10}
\end{figure*}

\begin{table}[htbp]
    \caption{Average R\textsuperscript{2} Score for Proposed Model For Training Dataset}
    \begin{center}
    \begin{tabular}{|c|c|c|c|}
    \hline
         \multirow{2}{*}{\textbf{Non-Dimensional Post Diameter ($F$)}} & \multicolumn{3}{c|}{\textbf{Average R\textsuperscript{2} Over $N$}}\\
         \cline{2-4}
         & \textbf{$u$} & \textbf{$v$} & \textbf{$p$} \\
         \hline
         0.25 & 0.988 & 0.999 & 0.816 \\
         0.30 & 0.999 & 0.999 & 0.812 \\
         0.35 & 0.999 & 0.999 & 0.812 \\
         0.40 & 0.999 & 0.999 & 0.799 \\
         0.45 & 0.999 & 0.999 & 0.799 \\
         0.50 & 0.999 & 0.999 & 0.834 \\
         0.55 & 0.999 & 0.999 & 0.820 \\
         0.60 & 0.999 & 0.999 & 0.877 \\
         0.65 & 0.999 & 0.999 & 0.885 \\
         0.70 & 0.999 & 0.999 & 0.896 \\
         \hline
         \textbf{Total} & \textbf{0.999} & \textbf{0.999} & \textbf{0.835} \\
    \hline
    \end{tabular}
    \label{tab:FieldR2Train}
    \end{center}
\end{table}

\begin{table}[htbp]
    \caption{Average R\textsuperscript{2} Score for Proposed Model For For Validation Dataset}
    \begin{center}
    \begin{tabular}{|c|c|c|c|}
    \hline
         \multirow{2}{*}{\textbf{Non-Dimensional Post Diameter ($F$)}} & \multicolumn{3}{c|}{\textbf{Average R\textsuperscript{2} Over $N$}}\\
         \cline{2-4}
         & \textbf{$u$} & \textbf{$v$} & \textbf{$p$} \\
         \hline
         0.21 & 0.988 & 0.999 & 0.826 \\
         0.32 & 0.999 & 0.999 & 0.820 \\
         0.33 & 0.999 & 0.999 & 0.818 \\
         0.34 & 0.999 & 0.999 & 0.814 \\
         0.43 & 0.999 & 0.999 & 0.798 \\
         0.47 & 0.999 & 0.999 & 0.800 \\
         0.49 & 0.999 & 0.999 & 0.804 \\
         0.51 & 0.999 & 0.999 & 0.810 \\
         0.52 & 0.999 & 0.999 & 0.815 \\
         0.57 & 0.999 & 0.999 & 0.845 \\
         0.62 & 0.999 & 0.999 & 0.866 \\
         0.68 & 0.999 & 0.999 & 0.888 \\
         \hline
         \textbf{Total} & \textbf{0.999} & \textbf{0.999} & \textbf{0.825} \\
    \hline
    \end{tabular}
    \label{tab:FieldR2Val}
    \end{center}
\end{table}

Fig.~\ref{fig:FieldPred_sup_F0_57_N10} compares predicted velocity and pressure distributions with CFD simulations for geometry $F = 0.57, N = 10$, which is a data point from validation dataset. The model successfully captures the overall flow structure across all fields, demonstrating accurate prediction of complex flow patterns around posts and through channels.

The error maps at the third column of Fig.~\ref{fig:FieldPred_sup_F0_57_N10} show numerical difference between predicted and simulated fields. The error magnitudes remain small across all fields.

The R² scores in Tables~\ref{tab:FieldR2Train}and~\ref{tab:FieldR2Val} show high scores in velocities near to 1. While pressure shows lower scores with 0.835 in training and 0.825 in validation, the pressure is only used for training physical patterns into the model and not used for particle trajectory prediction. Moreover, while not demonstrated in the figure, the CFD error exists in pressure fields inside the wall, showing sharp error rates, which reduces the R² scores as well. Thus, the model still demonstrates high performance in predicting fields.

\begin{figure}[htbp]
    \centering
    \subfigure[Bumped Mode Trajectory ($D_p$ = 0.0470)]{%
        \includegraphics[width=0.45\linewidth]{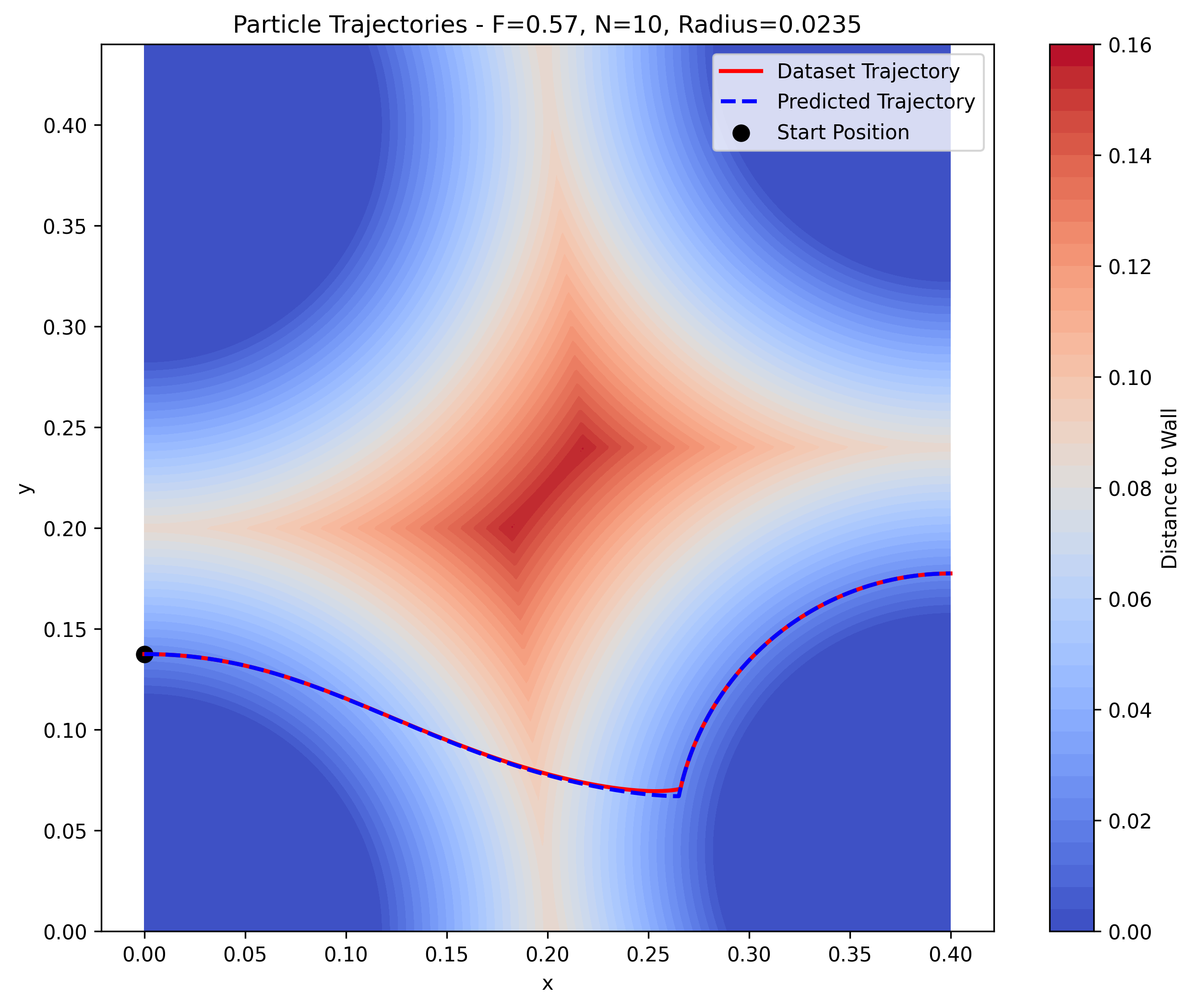}
        \label{fig:TrajectoryPred_sup_F0_57_N10_zigzag}
    }
    \hfill
    \subfigure[Zig-zag Mode Trajectory ($D_p$ = 0.0464)]{%
        \includegraphics[width=0.45\linewidth]{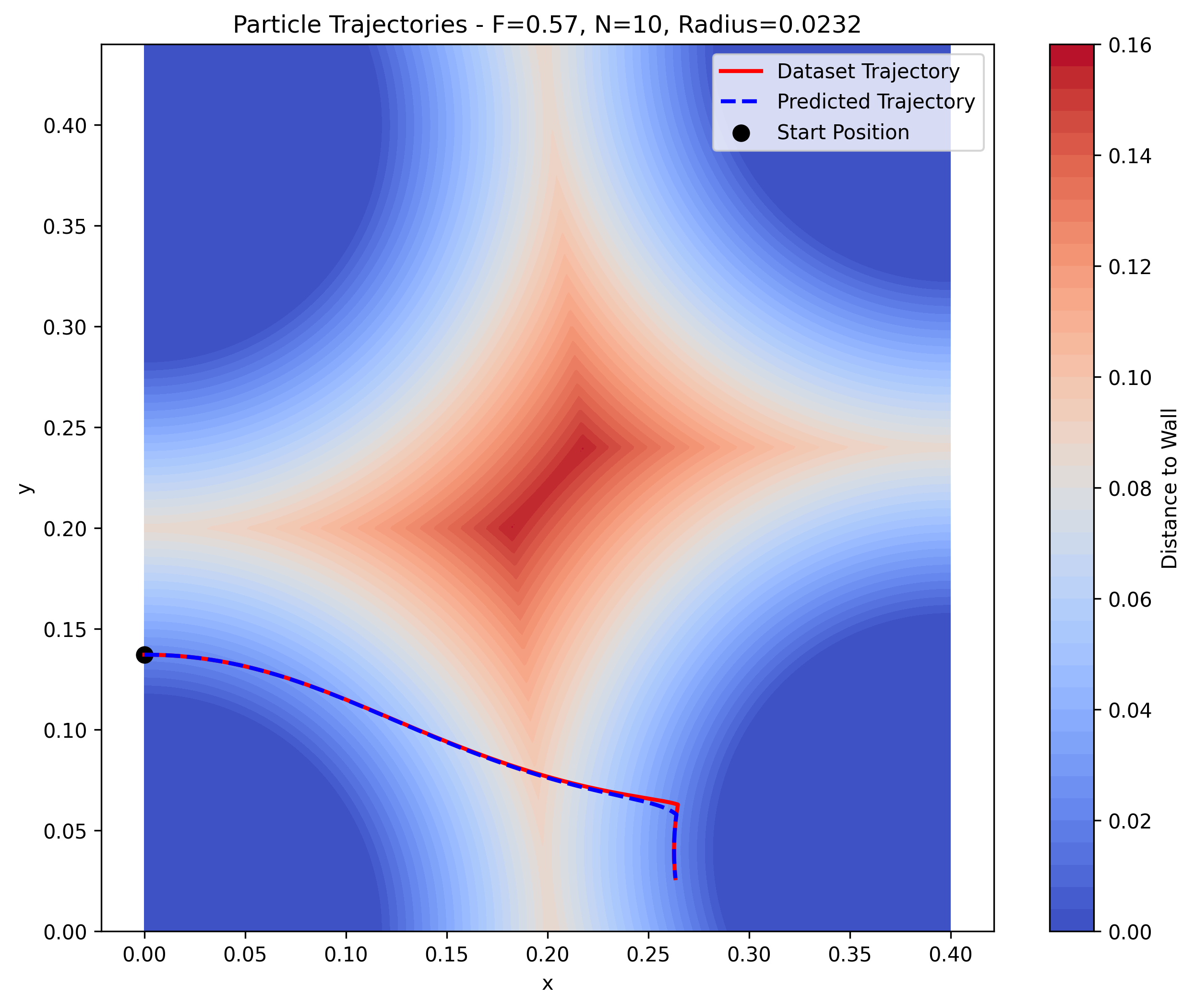}
        \label{fig:TrajectoryPred_sup_F0_57_N10_bumped}
    }
    \caption{Trajectory Prediction Comparison From Validation Data for F = 0.57 and N = 10 where $D_{c,CFD}$ = 0.0466}
    \label{fig:trajectory_comparison_F057_N10}
\end{figure}

\begin{figure}[htbp]
    \centering
    \subfigure[Bumped Mode Trajectory ($D_p$ = 0.0708)]{%
        \includegraphics[width=0.45\linewidth]{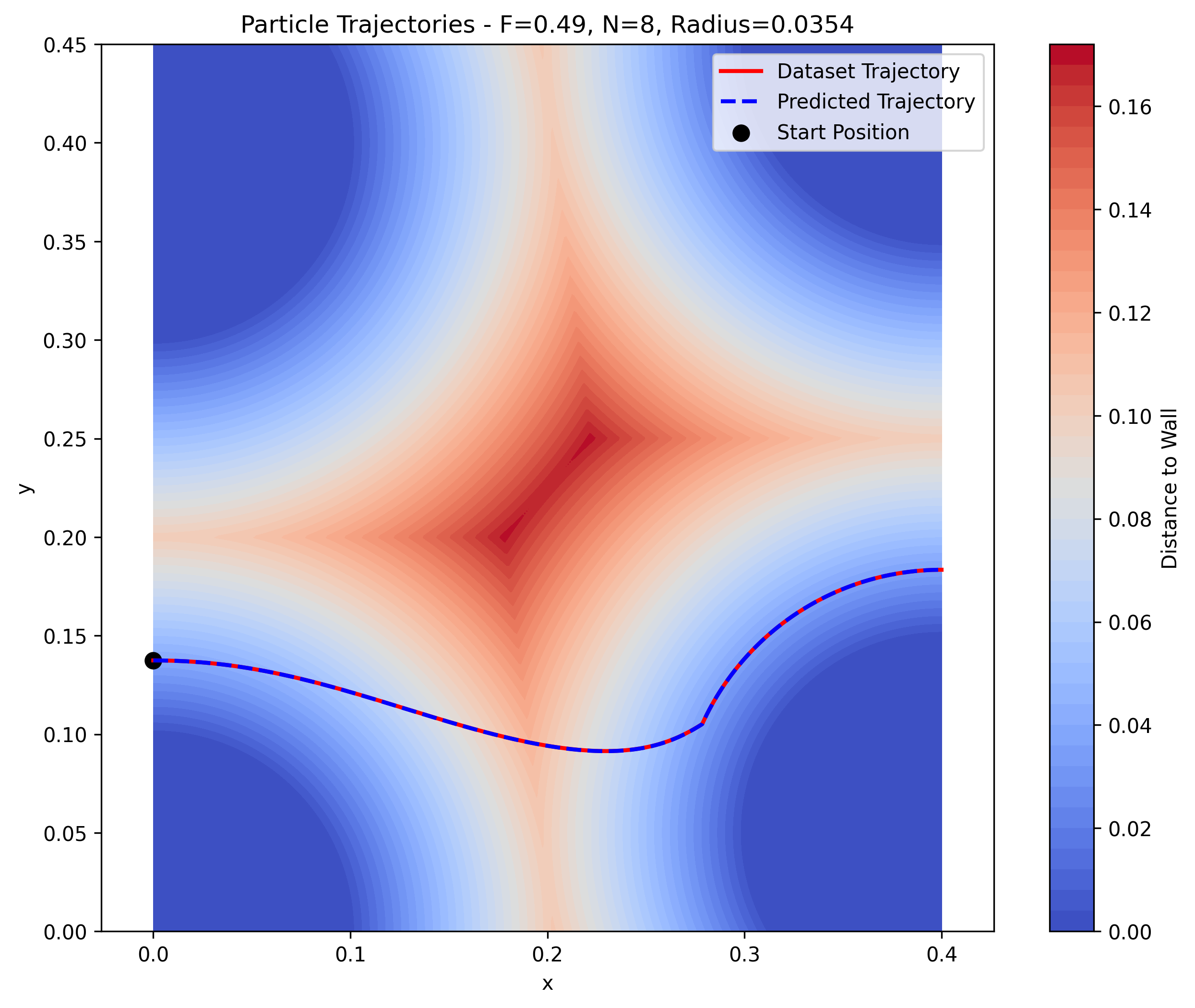}
        \label{fig:TrajectoryPred_sup_F0_49_N8_bumped}
    }
    \hfill
    \subfigure[Zig-zag Mode Trajectory ($D_p$ = 0.0700)]{%
        \includegraphics[width=0.45\linewidth]{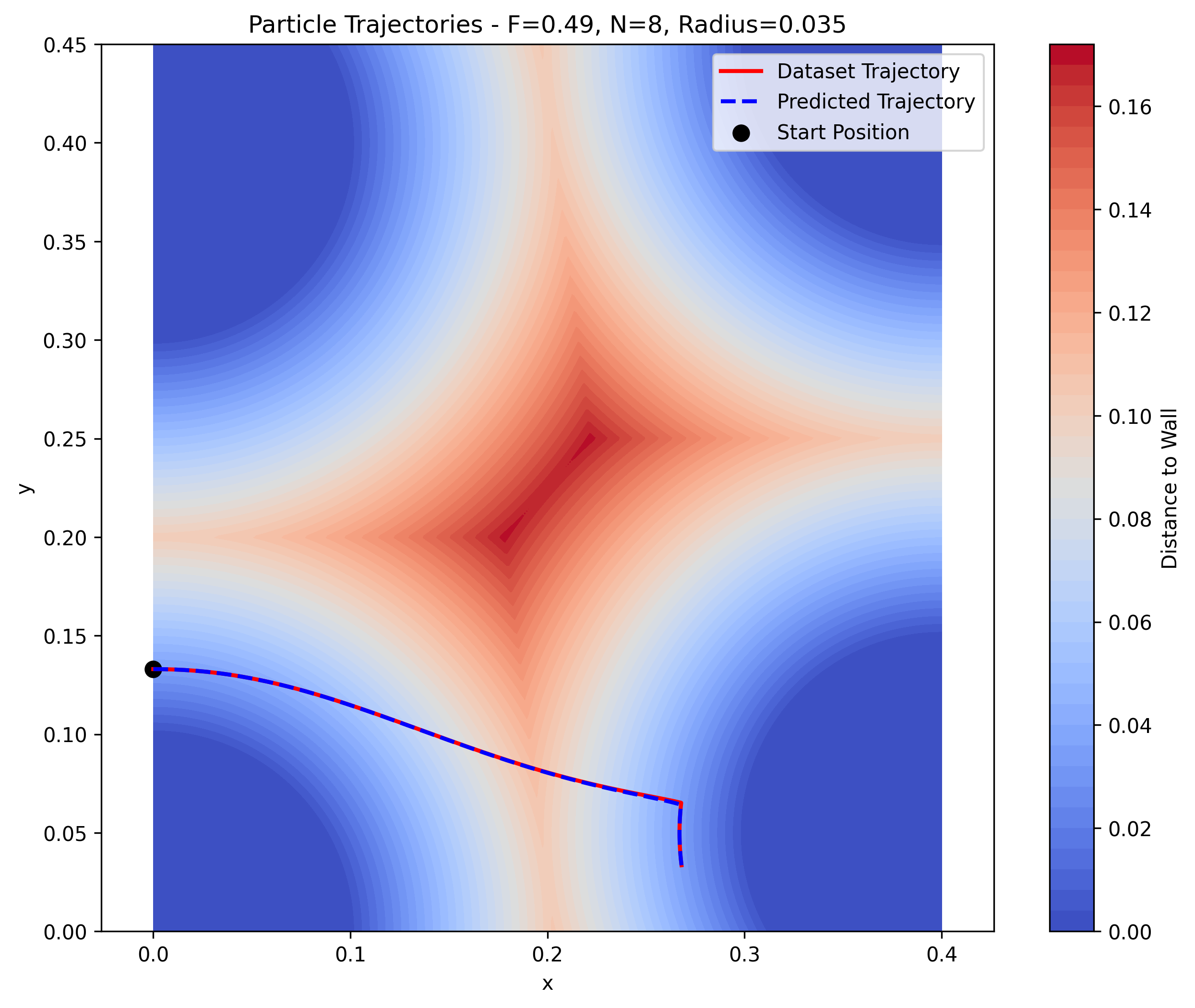}
        \label{fig:TrajectoryPred_sup_F0_49_N8_zigzag}
    }    
    \caption{Trajectory Prediction Comparison From Validation Data for F = 0.49 and N = 8 where $D_{c,CFD}$ =0 .0705}
    \label{fig:trajectory_comparison_F049_N8}
\end{figure}

\begin{figure}[htbp]
    \centering
    \subfigure[Bumped Mode Trajectory ($D_p$ = 0.0364)]{%
        \includegraphics[width=0.45\linewidth]{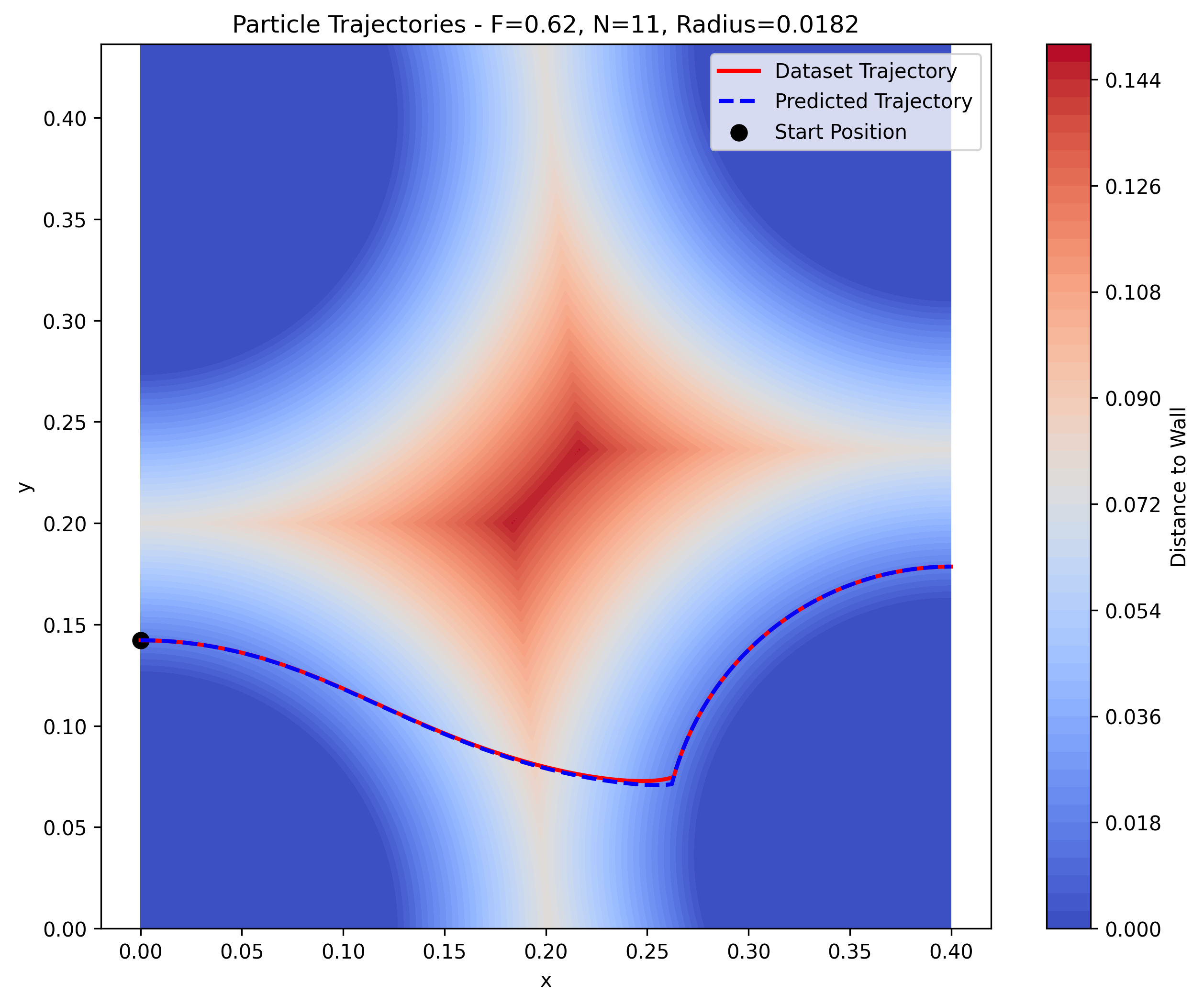}
        \label{fig:TrajectoryPred_sup_F0_62_N11_bumped}
    }
    \hfill
    \subfigure[Zig-zag Mode Trajectory ($D_p$ = 0.0360)]{%
        \includegraphics[width=0.45\linewidth]{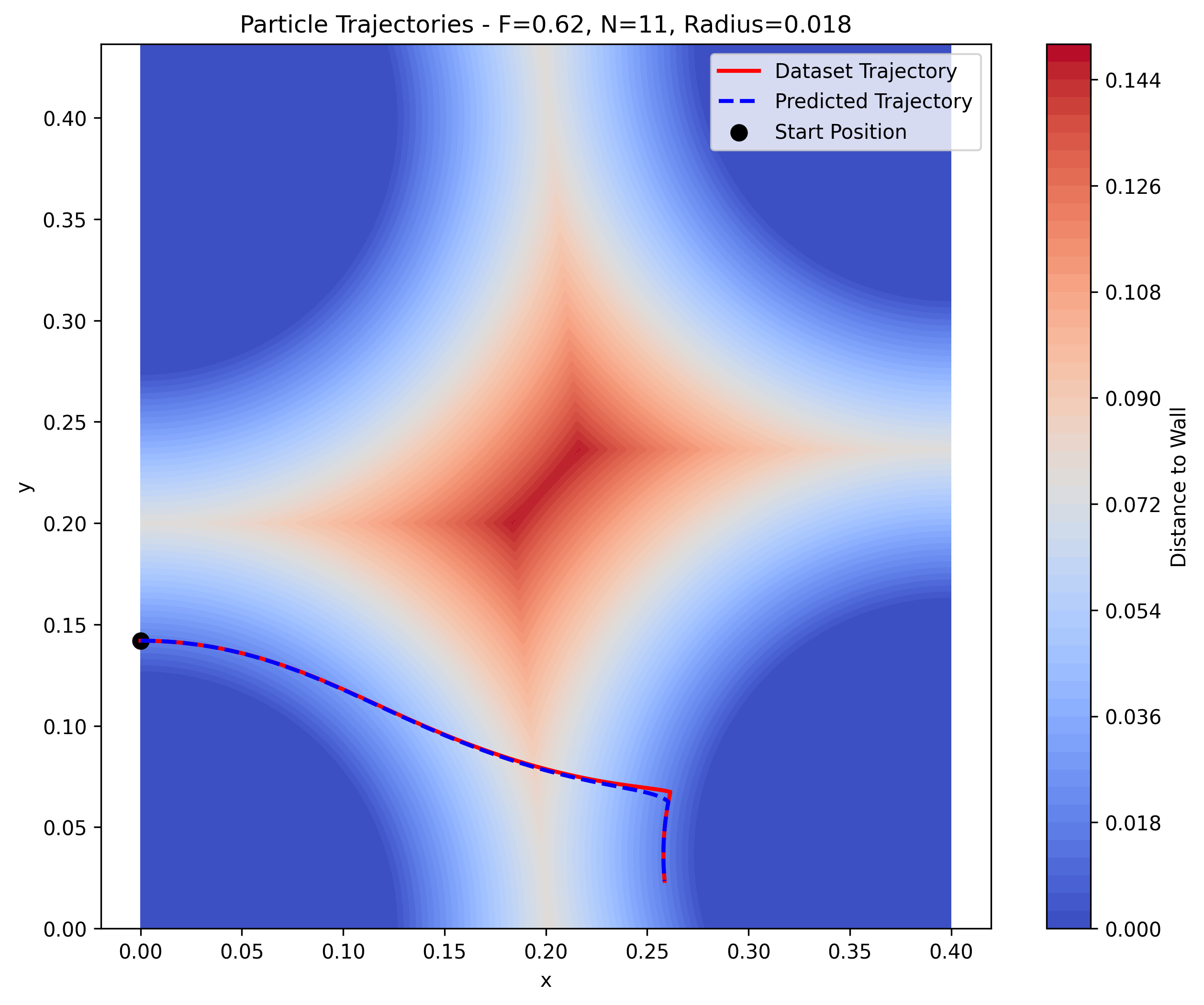}
        \label{fig:TrajectoryPred_sup_F0_62_N11_zigzag}
    }
    \caption{Trajectory Prediction Comparison From Validation Data for F = 0.62 and N = 11 where $D_{c,CFD}$ = 0.0361}
    \label{fig:trajectory_comparison_F062_N11}
\end{figure}

Particle trajectory validations in Fig.~\ref{fig:trajectory_comparison_F057_N10} confirm that predicted streamlines accurately follow CFD-computed paths. Fig.~\ref{fig:TrajectoryPred_sup_F0_57_N10_bumped} successfully reproduces bumped mode behavior when particle diameter ($D_p$) exceeds $D_c$ with $D_p$ = 0.0470 where $D_c$ = 0.466 from CFD simulation. Meanwhile, Fig.~\ref{fig:TrajectoryPred_sup_F0_57_N10_zigzag} successfully reproduces zig-zag mode behavior when $D_p$ is below $D_c$ with $D_p$ = 0.0464. The other plots (Figs.~\ref{fig:trajectory_comparison_F049_N8} and~\ref{fig:trajectory_comparison_F062_N11}) verify the performance in different geometric configurations as well.

\subsection{Critical Diameter Predictions}

\begin{table}[htbp]
    \caption{Average Critical Diameter Error for Baseline Model For Training Dataset}
    \begin{center}
    \begin{tabular}{|c|c|}
    \hline
         \textbf{Non-Dimensional Post Diameter ($F$)} & \textbf{Average Error \%}\\
         \hline
         0.25 & 0.792 \\
         0.30 & 1.748 \\
         0.35 & 2.297 \\
         0.40 & 3.066 \\
         0.45 & 3.995 \\
         0.50 & 4.580 \\
         0.55 & 4.469 \\
         0.60 & 3.030 \\
         0.65 & 3.134 \\
         0.70 & 5.580\\
         \hline
         \textbf{Total} & \textbf{3.269} \\
    \hline
    \end{tabular}
    \label{tab:DCErrorBase}
    \end{center}
\end{table}

\begin{table}[htbp]
    \caption{Average Critical Diameter Error for Proposed Model For Validation Dataset}
    \begin{center}
    \begin{tabular}{|c|c|}
    \hline
         \textbf{Non-Dimensional Post Diameter ($F$)} & \textbf{Average Error \%}\\
         \hline
         0.25 & 0.114 \\
         0.30 & 0.064 \\
         0.35 & 0.106 \\
         0.40 & 0.126 \\
         0.45 & 0.182 \\
         0.50 & 0.207 \\
         0.55 & 0.361 \\
         0.60 & 0.449 \\
         0.65 & 0.721 \\
         0.70 & 1.011\\
         \hline
         \textbf{Total} & \textbf{0.334} \\
    \hline
    \end{tabular}
    \label{tab:DCErrorSupervisedTrain}
    \end{center}
\end{table}

\begin{table}[htbp]
    \caption{Average Critical Diameter Error for Proposed Model For Validation Dataset}
    \begin{center}
    \begin{tabular}{|c|c|}
    \hline
         \textbf{Non-Dimensional Post Diameter ($F$)} & \textbf{Average Error \%}\\
         \hline
         0.21 & 2.479 \\
         0.32 & 0.095 \\
         0.33 & 0.090 \\
         0.34 & 0.084 \\
         0.43 & 0.115 \\
         0.47 & 0.190 \\
         0.49 & 0.201 \\
         0.51 & 0.273 \\
         0.52 & 0.353 \\
         0.57 & 0.447\\
         0.62 & 0.592 \\
         0.68 & 0.815\\
         \hline
         \textbf{Total} & \textbf{0.478} \\
    \hline
    \end{tabular}
    \label{tab:DCErrorSupervisedVal}
    \end{center}
\end{table}

Tables~\ref{tab:DCErrorBase} and~\ref{tab:DCErrorSupervisedTrain} compare critical diameter prediction accuracy between baseline and proposed models. Both approaches show increasing error with higher post diameters, reaching maximum error at $F = 0.7$.

The proposed model achieves a total average error of 0.334\%, representing an 89.8\% improvement over the baseline model's 3.269\% error in training dataset. This substantial improvement demonstrates the effectiveness of the enhanced network architecture and periodic layer integration.

In validation dataset shown in Table~\ref{tab:DCErrorSupervisedVal}, the error increased to 0.478\%, which still represents 85.4\% improvement over baseline model's error in training dataset. Meanwhile, the model showed sharp rise in error in F = 0.21 as F is outside of training range of 0.25 to 0.7. 

\subsection{Ablation Study - With/Without Periodic Layer Comparison}

To isolate the effects of periodic layer enforcement, we trained a variant without periodic layers, instead using soft periodic loss terms during training while maintaining identical architecture and training procedures otherwise. \\

\subsubsection{Critical Diameter Error}

\begin{table}[htbp]
    \caption{Average Critical Diameter Error for Proposed Model without Periodic Layer For Training Dataset}
    \begin{center}
    \begin{tabular}{|c|c|}
    \hline
         \textbf{Non-Dimensional Post Diameter ($F$)} & \textbf{Average Error \%}\\
         \hline
         0.25 & 0.027 \\
         0.30 & 0.066 \\
         0.35 & 0.032 \\
         0.40 & 0.058 \\
         0.45 & 0.074 \\
         0.50 & 0.126 \\
         0.55 & 0.128 \\
         0.60 & 0.241 \\
         0.65 & 0.519 \\
         0.70 & 0.454 \\
         \hline
         \textbf{Total} & \textbf{0.172} \\
    \hline
    \end{tabular}
    \label{tab:DCErrorNoPeriodic_Train}
    \end{center}
\end{table}

\begin{table}[htbp]
    \caption{Average Critical Diameter Error for Proposed Model without Periodic Layer For Validation Dataset}
    \begin{center}
    \begin{tabular}{|c|c|}
    \hline
         \textbf{Non-Dimensional Post Diameter ($F$)} & \textbf{Average Error \%}\\
         \hline
         0.21 & 0.250 \\
         0.32 & 0.055 \\
         0.33 & 0.047 \\
         0.34 & 0.041 \\
         0.43 & 0.083 \\
         0.47 & 0.075 \\
         0.49 & 0.085 \\
         0.51 & 0.123 \\
         0.52 & 0.102 \\
         0.57 & 0.184 \\
         0.62 & 0.410 \\
         0.68 & 0.455 \\
         \hline
         \textbf{Total} & \textbf{0.159} \\
    \hline
    \end{tabular}
    \label{tab:DCErrorNoPeriodic_Val}
    \end{center}
\end{table}

Tables~\ref{tab:DCErrorNoPeriodic_Train} and~\ref{tab:DCErrorNoPeriodic_Val} show that the model without periodic layers achieves 0.172\% average error on training and 0.159\% on validation, lower than the periodic layer implementation. This improvement likely results from unified Swish activation across all layers, whereas the periodic layer model uses tanh activation in the periodic layer. \\

\subsubsection{Periodicity Consistency}

\begin{figure}[htbp]
    \centering
    \includegraphics[width=0.8\linewidth]{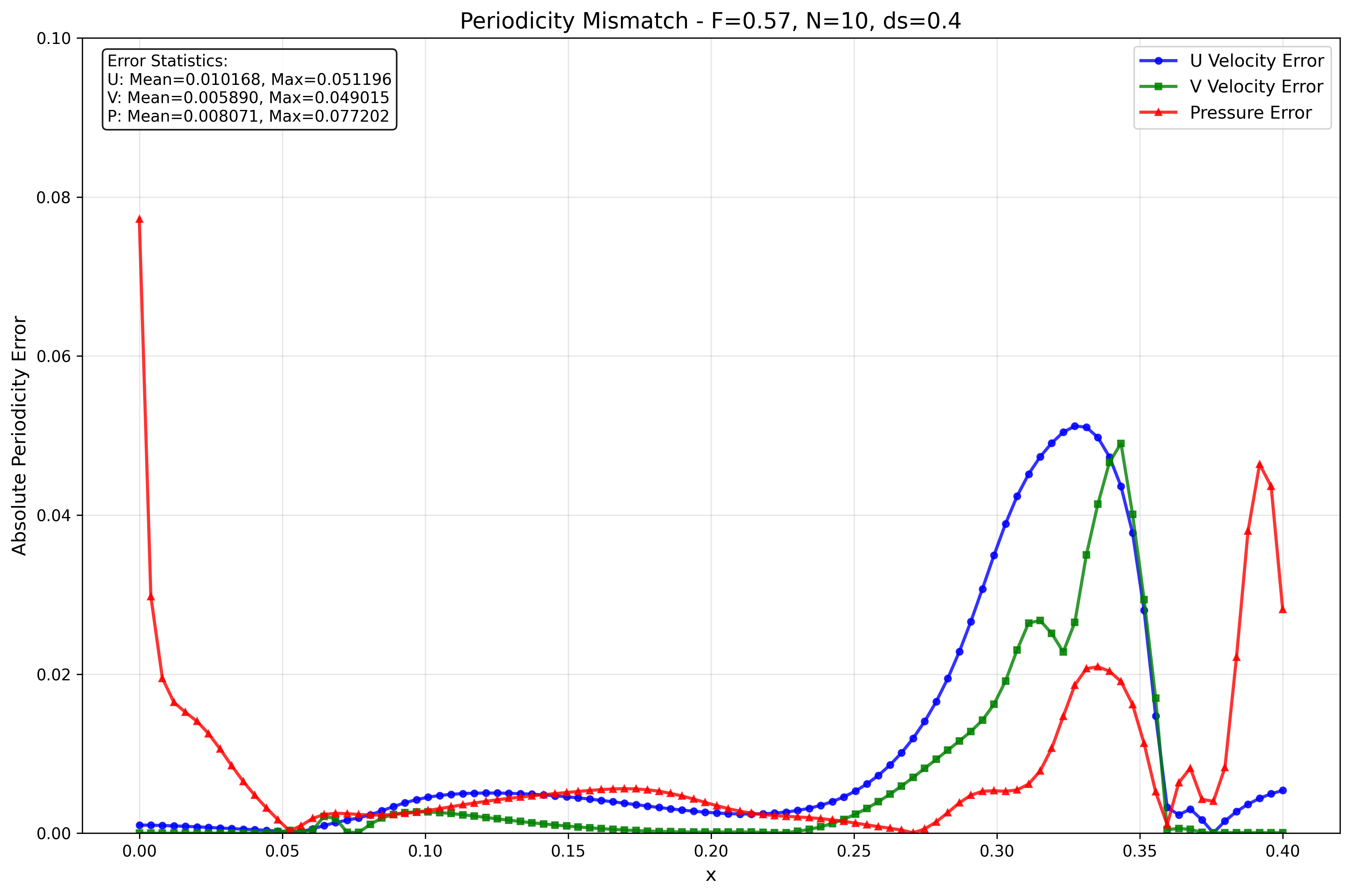}
    \caption{Periodicity Error Across x for F = 0.57 and N = 10 without Periodic Layers}
    \label{fig:PeriodicityError}
\end{figure}

Fig.~\ref{fig:PeriodicityError} evaluates periodicity enforcement across the x-direction by measuring differences in $(u,v,p)$ between top and bottom boundaries. The model with periodic layers achieves exact matching for all three fields (zero error). However, the soft enforcement approach exhibits significant periodicity violations. The average errors are 0.01017 for u, 0.00589 for v, 0.00807 for p. The max error is 0.05120 for u, 0.04902 for v, and 0.07720 for p. These errors show periodicity of DLD device is not preserved.

Thus, although the model without periodic layers shows slightly lower critical diameter error, the periodicity mismatch represents a critical limitation that could accumulate across multiple unit cells in practical DLD devices, potentially leading to substantial trajectory deviations in full-scale implementations.


\section{Discussion}

\subsection{Importance of Periodicity in DLD Device Design}

Exact periodic boundary conditions are critical for DLD device modeling because surrogate models predict only single unit cells representing segments of longer post arrays. While critical diameter errors affect individual calculations without accumulation, periodicity mismatches propagate across multiple unit cells, leading to cumulative deviations in CTC trajectory predictions.

The ablation study reveals a fundamental trade-off: models without periodic layers achieve slightly lower critical diameter error (0.172\% vs 0.478\%) but exhibit significant periodicity violations. For practical DLD applications requiring multiple unit cells, exact periodicity enforcement provides superior stability and reliability despite marginally higher single-unit errors. This demonstrates that architectural solutions for boundary conditions offer advantages over soft enforcement approaches in multi-unit systems.

\subsection{Advantages of Field-Based Surrogate Modeling}

Previous DLD surrogate models focused on direct particle trajectory prediction \cite{chen2024poster}, which, while computationally efficient, limits generalization and physics-informed development by bypassing underlying fluid mechanics. Our field-based approach predicting complete velocity and pressure distributions offers several key advantages:

\textbf{-Enhanced Analysis Capabilities:} Complete flow field descriptions enable post-hoc optimization and detailed flow analysis without additional simulations, supporting iterative design processes.

\textbf{-Gradient-Based Optimization:} Differentiable field predictions facilitate gradient-based geometric parameter optimization for future automated design workflows.

\textbf{-Decoupled Particle Physics:} Flow fields separate from particle-specific calculations allow analysis of different particle sizes and shapes without model retraining, enhancing versatility for various cancer cell types.

Our results demonstrate that accurate field predictions naturally translate to precise critical diameter estimations and trajectory calculations, validating this approach's effectiveness.

\subsection{Dataset Selection and Future Training Approaches}

In this study, we employed a pressure-based dataset where pressure difference ($\Delta p$) is specified at the inlet and outlet boundaries (Eqn.~\ref{dp}), allowing the flow rate and velocity profile to develop naturally from the geometry. This choice was made with consideration for future label-free training approaches. Pressure-based boundary conditions are more convenient for Physics-Informed Neural Network (PINN) implementations, as they provide more straightforward constraints for incorporating Navier-Stokes residuals into the loss function without requiring specification of complete velocity profiles at boundaries.

While pressure-based datasets offer advantages for physics-informed training, Reynolds number-based datasets could also be employed for DLD device modeling. Reynolds number-based approaches would explicitly incorporate $Re$ as an input parameter, enabling the model to learn flow behavior across different flow regimes. The selection between these approaches depends on the specific application requirements and whether the focus is on geometric optimization at fixed flow conditions or comprehensive flow regime analysis.

\subsection{Limitations}

Despite significant improvements, our model faces several constraints. Requires large size training data, and training time remains substantial at 6 hours per model, potentially limiting rapid parameter exploration across multiple geometric variations. Moreover, The current validation focuses on limited geometric parameters ($D_0$ and $N$) while excluding post shapes, fluid properties, and Reynolds numbers, restricting generalization to broader DLD design spaces.

\subsection{Future Work}

\subsubsection{Label Free Training with Navier-Stokes PINN Residuals}

A promising future direction is to incorporate physics-informed loss terms directly from the Navier-Stokes equations to remove the reliance on CFD-generated labels. Such label-free training has been done on simple structure such as circular pipe or blood vessels by Sun et al.. \cite{Sun_2020} This approach would leverage the governing equations to regularize the network and reduce data requirements. From currently proposed model, the governing equations and flow direction enforcements must be driven to lead to label-free training in DLD devices.

\subsubsection{Geometric Generalization}

Model extension to broader DLD parameter spaces requires incorporating variable viscosity, Reynolds numbers, post shapes, and irregular gap configurations. This generalization would enable surrogate modeling for diverse DLD applications beyond the current fixed-parameter validation.

Variable Reynolds number modeling presents particular interest for different fluid conditions and flow regimes relevant to various biological sample types. Post shape variations (elliptical, square, etc.) could optimize separation efficiency for specific cell populations, while irregular gap patterns may enhance separation resolution through engineered flow asymmetries.


\section{Conclusion}
This paper introduced the Periodicity-Enforced Neural Network, a novel approach for designing Deterministic Lateral Displacement devices that enforces exact periodicity with computational efficiency. The proposed method addresses key limitations of existing surrogate models by incorporating periodic layers that guarantee boundary consistency.

The model validation demonstrated the effectiveness of the approach, achieving a critical diameter prediction error of only 0.478\%, representing an 85.4\% improvement over baseline methods. The periodic layer implementation ensured exact matching of velocity and pressure fields at unit cell boundaries, addressing a critical requirement for multi-unit DLD device modeling that soft enforcement approaches cannot guarantee.

The proposed framework offers several significant advantages for DLD device design: (1) the field-based prediction approach provides complete flow descriptions enabling further analysis and optimization, (2) exact periodicity enforcement ensures stability across multiple unit cells, and (3) the machine learned models enable generalization to new geometric configurations without retraining.

While the current implementation focuses on specific geometric parameters, the framework provides a foundation for more comprehensive DLD device design tools. The demonstrated ability to train surrogates to replace CFD simulations with periodicity enforced is valuable for rapid prototyping and optimization of microfluidic devices for cancer detection applications.

The success of this approach bridges to future possibilities for physics-informed surrogate modeling in microfluidics, where computational efficiency and prediction performance is achieved through dataless training. Future extensions incorporating physics-informed losses and broader geometric generalization promise to further enhance the capabilities of this innovative design methodology.


\bibliographystyle{IEEEtran}   
\bibliography{references}            

\end{document}